\documentclass[11pt]{article}
\usepackage{coling2020}
\usepackage{times}
\usepackage{url}
\usepackage{latexsym}

\usepackage{graphicx}
\usepackage{caption}
\usepackage{subcaption}
\usepackage{amsmath,amssymb}

\colingfinalcopy

\title{Language-Driven Region Pointer Advancement \\ for Controllable Image Captioning}

\author{Annika Lindh \and Robert Ross  \and John D. Kelleher \\
  ADAPT Research Centre \\
  School of Computer Science,\\
  Technological University Dublin, \\ 
  Kevin Street, Dublin 8, IRELAND \\
  {\tt  \{annika.lindh;robert.ross;john.d.kelleher\}@tudublin.ie }\\}

\date{}

\begin{document}
\maketitle

\begin{abstract}
Controllable Image Captioning is a recent sub-field in the multi-modal task of Image Captioning wherein constraints are placed on which regions in an image should be described in the generated natural language caption. This puts a stronger focus on producing more detailed descriptions, and opens the door for more end-user control over results. A vital component of the Controllable Image Captioning architecture is the mechanism that decides the timing of attending to each region through the advancement of a region pointer. In this paper, we propose a novel method for predicting the timing of region pointer advancement by treating the advancement step as a natural part of the language structure via a NEXT-token, motivated by a strong correlation to the sentence structure in the training data. We find that our timing agrees with the ground-truth timing in the Flickr30k Entities test data with a precision of 86.55\% and a recall of 97.92\%. Our model implementing this technique improves the state-of-the-art on standard captioning metrics while additionally demonstrating a considerably larger effective vocabulary size.
\end{abstract}

\section{Introduction}

\blfootnote{    
     \hspace{-0.65cm} 
     This work is licensed under a Creative Commons 
     Attribution 4.0 International License.
     License details:
     \url{http://creativecommons.org/licenses/by/4.0/}.
}

Image Captioning brings together the two fields of Computer Vision and Natural Language Generation into a task where the model needs to translate an input image into an appropriate natural language text description. The task leaves some ambiguity regarding which parts of the image should be mentioned and which ones can be excluded. This has led to a common problem where models tend to generate overly generic descriptions that seem to focus more on the category of the image than on its individual content \cite{quirks,image_role}.

\begin{figure}
  \centering
   \includegraphics[height=6cm]{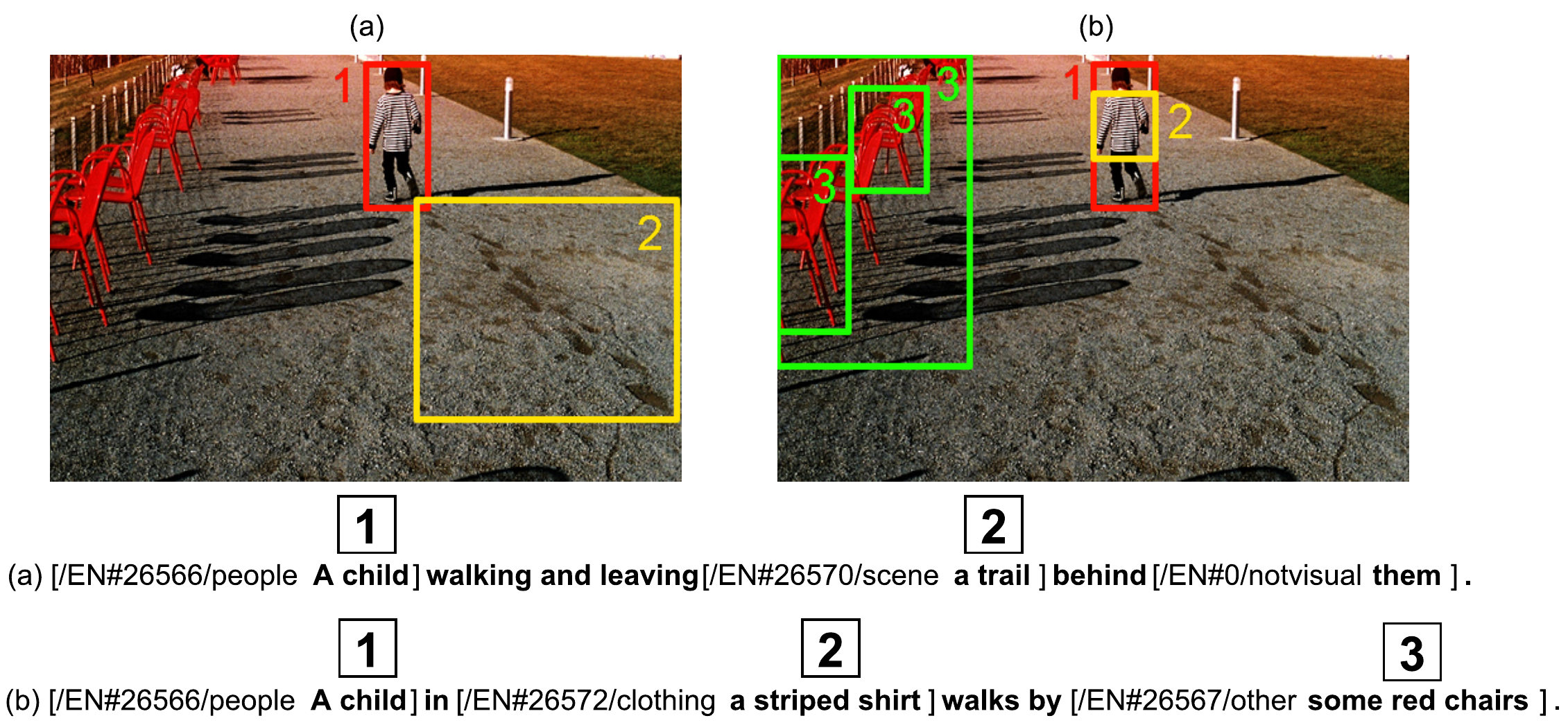}
  \caption{Image and captions from Flickr30k \cite{flickr30k} with regions and text annotations from Flickr30k Entities \cite{flickr30k_entities}. Square brackets indicate entity annotations. Some entity regions consist of multiple bounding boxes.}
   \label{fig_data}
\end{figure}

Recently, \newcite{sct} introduced Controllable Image Captioning as a new sub-task of Image Captioning with a stronger focus on image details. In this task, the input to the model is an image along with bounding box coordinates for a sequence of regions (where each region can consist of one or more bounding boxes) that must be explicitly described in the candidate caption. Thus, in contrast to standard Image Captioning, a generic candidate caption would not meet the criteria of a suitable caption even if it contained no factual errors; this is reflected in the evaluation process where the candidate caption is only compared to those ground-captions that share the same sequence of regions. Fig.~\ref{fig_data} shows an image and two corresponding captions from the Flickr30k dataset \cite{flickr30k} along with complementary data from Flickr30k Entities \cite{flickr30k_entities} which provides annotations that link entities in the captions to region bounding boxes in the corresponding image.

In terms of practical use, a Controllable Image Captioning model provides more flexibility and user-control over the generated captions, without having to retrain the model. Since the region selection process is not entangled with the caption generation components, the former can be swapped out to adapt to different scenarios (e.g. when applied to social media images, it could target regions where a facial recognition system has tagged friends of the user). Furthermore, it opens up the possibility of adhering to individual user preferences regarding the amount and type of details to describe -- this is a feature that could provide real benefits to blind and low-vision users \cite{blv}.

When it comes to model architecture, a key component in Controllable Image Captioning is the mechanism that advances the region pointer at the appropriate time so that each requested region is sufficiently described. In previous work, this has been approached as its own prediction task in parallel to the prediction of the next word \cite{sct}. However, in this paper, we demonstrate that the timing of the region pointer advancement is strongly linked to the sentence structure in the ground-truth and thus could likely be predicted as if it were a natural language token. Hence, we propose a novel approach to the region pointer advancement mechanism that directly leverages the language model to generate a unique NEXT-token as part of the caption generation process. We implement this in a Controllable Image Captioning model where we demonstrate its effectiveness by measuring how often the NEXT-token is predicted in agreement with the ground-truth. Furthermore, we measure the model's overall performance on standard Image Captioning metrics where it outperforms the current state-of-the art on the Flickr30k \cite{flickr30k} benchmark test. Additionally, we observe that our model demonstrates a considerably larger effective vocabulary size than the current state of the art model, thus enabling our model to describe a greater variety of visual content.

We make our code and trained model publicly available for future work to build on.\footnote{\url{https://github.com/AnnikaLindh/Controllable_Region_Pointer_Advancement}}

\section{Related Work} \label{sec_related}
Controllable Image Captioning builds directly on the standard Image Captioning task and thus involves a similar set of components, with the main difference being the role of visual attention. In the early neural models for Image Captioning, visual information was typically extracted into a fixed-size vector representation of the full image, which would be provided as input either at the start or at each recurrent timestep in the generation process (e.g. \newcite{babytalk}, \newcite{fang_2015}, \newcite{show_tell}). These early neural models were found to produce largely generic captions with output similar to retrieval-based models \cite{quirks,image_role}.

In recent years, dynamically generated attention has become the standard method to mitigate generic captions through enhanced visual grounding by providing the model with only the visual information relevant to the current timestep; multiple variations of dynamic attention have been explored, from the soft and hard attention proposed by \newcite{sat} to the region proposal network used by \newcite{bottom_up}.

In Controllable Image Captioning, the locations of the regions of interest are included in the input to the model (as shown in Fig.~\ref{fig_data}). To extract meaningful features from these regions, previous work in Controllable Image Captioning has successfully used the region features from \newcite{bottom_up} that were developed for the standard captioning task but whose object-focused training method corresponds well to the object-focused region annotations in the Flickr30k Entities dataset \cite{flickr30k_entities}. 

Since the \textit{selection and ordering} of the regions is fixed, the challenge instead lies in predicting the appropriate \textit{timing} of attending to each region. Since the current task requires that the output incorporates exactly the requested regions, the model must ensure that each part of the generated caption is sufficiently grounded in its corresponding visual region and that the generated sequence of words is not terminated before the complete region sequence has been described.

To predict the timing of attending to each region, \newcite{sct} implement a region pointer along with a mechanism to predict, at each timestep, whether this pointer should be incremented or not. The prediction follows a multi-step process that includes an additional LSTM layer to model the \textit{attention state} of the current caption chunk, extended with an additional output layer to model the \textit{end} of this chunk, as well as a chunk-shifting gate module that outputs the probability of incrementing the region pointer, based on: the chunk's attention state, the chunk's end state and the visual features for each of the requested regions. A potential weakness in this implementation is the structure of connection between the word prediction LSTM, the attention LSTM and the sampling of the next word: the chunk-shifting gate (which comprises an earlier step in this module) must make its decision \textit{before} the next word has been sampled. Thus, it must predict whether the next word \textit{is going to be} the last word of the chunk rather than predicting whether the current word \textit{is} the end of the chunk. This complicates the task of the chunk-shifting gate which must rely on incomplete information about the chunk when predicting its end-point. Furthermore, it would be difficult for the chunk-shifting gate to anticipate the result of any non-deterministic method for sampling the next word.

In the following section, we argue that a more elegant solution for advancing the region pointer is not only possible but also preferable.

\section{Region Pointer Advancement} \label{sec_pointer}
Region pointer advancement training relies on separating full captions into chunks that each relate to a visual region. We use the same chunking method as \newcite{sct} on the Flickr30k Entities \cite{flickr30k_entities} captions, meaning that each chunk starts with the word immediately following the end of the previous chunk, and ends with the last word in the following entity annotation that has at least one bounding box associated with it; entity annotations without bounding boxes are not taken into account.

Fig.~\ref{fig_data} shows two example captions with entity annotations enclosed in brackets, where each entity annotation includes an entity ID, an entity type and the noun phrase associated with that entity. After chunking, example \textit{a} consists of three chunks: 1) \textit{a child}, 2) \textit{walking and leaving a trail} and 3) \textit{behind them}. The visual region associated with each chunk refers to the average-pooled features of all bounding boxes associated with its entity. Since the third chunk does not have an associated entity, we associate it with the empty (zero-vector) region during training.

\begin{figure}
\centering
\begin{subfigure}{.5\textwidth}
  \centering
  \includegraphics[width=1\linewidth]{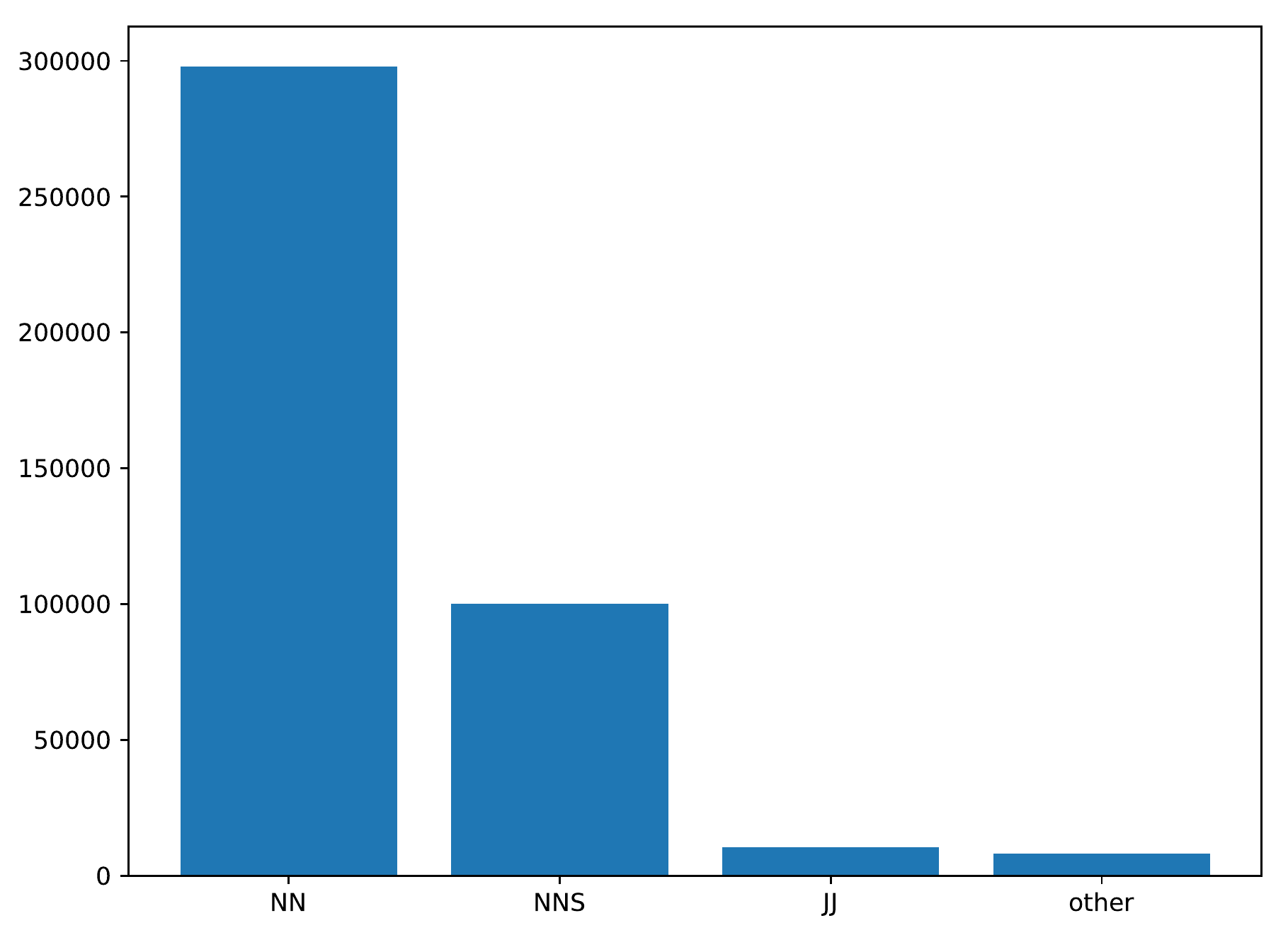}
  \caption{}
  \label{fig_eoc_freq}
\end{subfigure}%
\begin{subfigure}{.5\textwidth}
  \centering
  \includegraphics[width=0.96\linewidth]{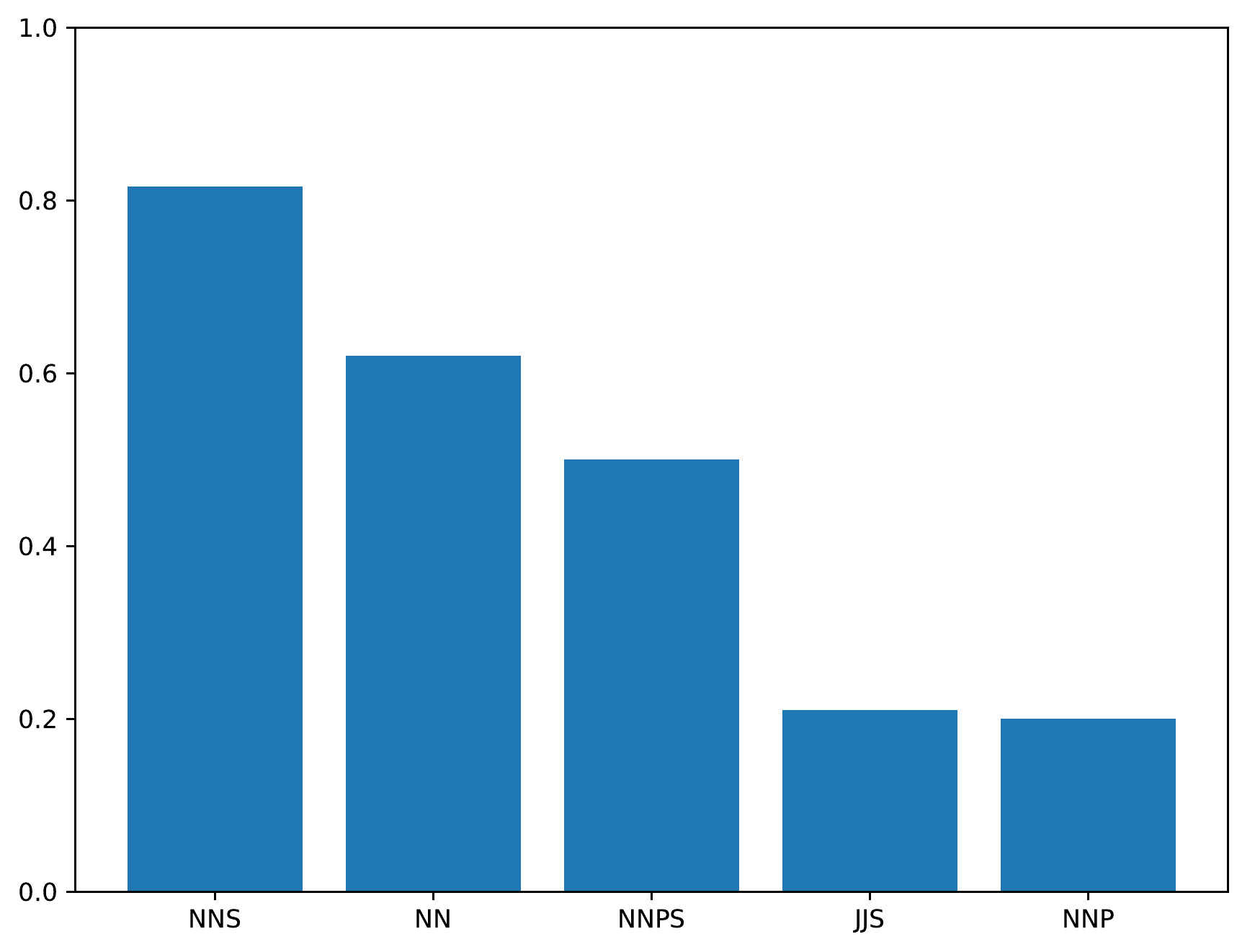}
  \caption{}
  \label{fig_eoc_pp}
\end{subfigure}
\caption{PoS tag statistics for the final word of chunks in the training data. (a) shows the total counts of each tag appearing at the end of a chunk, while (b) shows the positive predictive value for indicating the end of a chunk of the top five most predictive tags.}
\label{fig_eoc_stats}
\end{figure}

Since the end of a chunk is related to the position of its entity's noun phrase, it seems reasonable to suspect that the end of a chunk is correlated to parts-of-speech (PoS). We explore two potential correlations after employing automatic PoS tagging on the training data, using the Python implementation of the Brill tagger \cite{brill} found in the TextBlob\footnote{https://textblob.readthedocs.io} library. Fig.~\ref{fig_eoc_freq} shows the frequency of each PoS tag assigned to the last word of a chunk; unsurprisingly, the last word of a chunk is most commonly tagged as a noun. Further, Fig.~\ref{fig_eoc_pp} shows the positive predictive value (i.e. precision, which takes into account how often each PoS tag appears in the training data) for indicating the end of a chunk among the top five most predictive PoS tags. Again, nouns are most commonly associated with ending a chunk, though their positive predictive value varies from 81.6\% for plural nouns (NNS) to 20.0\% for singular proper nouns (NNP).

From the statistics shown in Fig.~\ref{fig_eoc_stats}, it is clear that the end of a chunk (and thus, the region pointer advancement timing) is correlated to the sentence structure. Thus, we propose to inject a special language token (NEXT) to mark the end-of-chunk events, treating it in a similar way to the beginning-of-sequence (BOS) and end-of-sequence (EOS) events. The two examples in Fig.~\ref{fig_data} thus become: (a) \textit{BOS a child NEXT walking and leaving a trail NEXT behind them EOS}, and (b) \textit{BOS a child NEXT in a striped shirt NEXT walks by some red chairs NEXT EOS}. With the added NEXT-tokens, our model no longer requires additional steps or learnable layers to predict the region advancement timing since this is treated as a natural language property and predicted by the language model (as any other word). Unlike the method used by \newcite{sct} (described in section~\ref{sec_related}), our prediction mechanism has access to the most recently generated word regardless of sampling method. Additionally, during the immediately following timestep, the NEXT-token becomes the previous word, and thus our language model is explicitly informed that a new chunk should begin.

To evaluate our proposed region pointer advancement method, we implement it as part of a Controllable Image Captioning model (described in section~\ref{sec_model}) and test it on the region sequence scenario \cite{sct} where each example's candidate caption is compared only to those ground-truth captions that share the same region sequence (i.e. the same regions in the same order).

\section{Model Architecture} \label{sec_model}

\begin{figure}
  \centering
  \includegraphics[height=3.4cm]{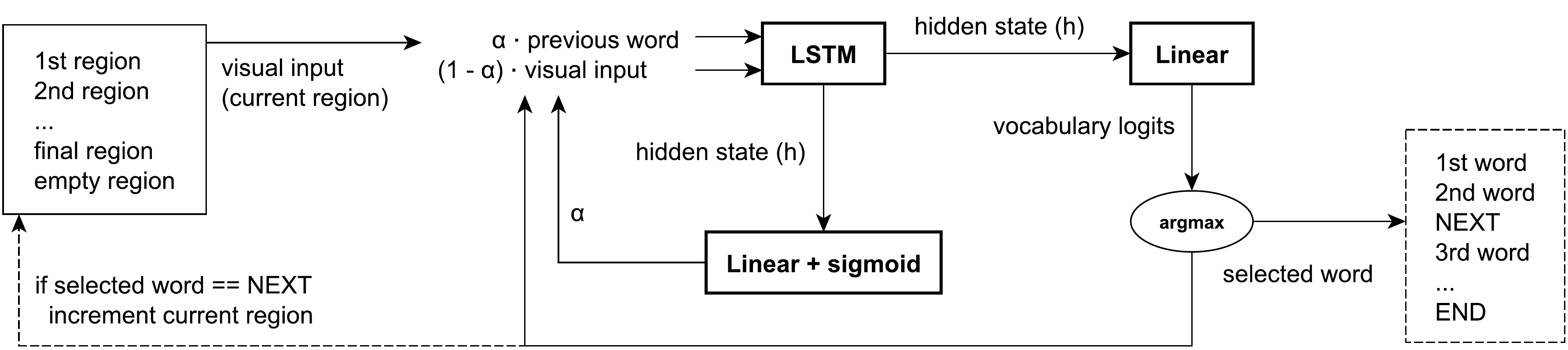}
  \caption{Model architecture and flowchart for recurrent steps. Solid line arrows indicate direct information flow, while the dashed line arrow indicates an action taken based on the information.}
   \label{fig_model_arch}
\end{figure}

Our architecture, shown in Fig.~\ref{fig_model_arch}, is similar to a typical Image Captioning model based on the encoder-decoder design with visual attention. The recurrent unit is a two-layer Long Short-Term Memory (LSTM) \cite{lstm} module with a hidden size of 1024 units (using the implementation details of \newcite{lstm_multi}). The input \(i_{t}\) at each timestep \(t\) is the weighted concatenation of the previous word embedding \(w_{t-1}\) and the current visual region's embedding \(v_{t}\), with the weight balance \(\alpha_{t}\) between them being generated according to equations~\ref{eq_linear_sigmoid}-\ref{eq_concat}:

\begin{align}
  \alpha_{t} & = \sigma ( h_{t-1} \cdot W_{\alpha} + b_{\alpha} ) \label{eq_linear_sigmoid} \\
  i_{t} & = [\alpha_{t} \cdot w_{t-1}, (1-\alpha_{t}) \cdot v_{t}] \label{eq_concat}
\end{align}

\noindent
where \(W_{\alpha}, b_{\alpha}\) are the learnable weights and bias of a single linear layer. The hidden state from the LSTM is passed through a single learnable linear layer to produce the logits over the full vocabulary.

Our word embeddings are randomly initialized learned vector representations of size 1024, while the visual embedding is a vector of size 2053. The first 2048 features are extracted from the final pooling layer (pool5) of a Bottom-Up network \cite{bottom_up} where the region's bounding box(es) are used as the region proposals; in the case where a region is made up of multiple bounding boxes, we average-pool their features into a single vector representation of size 2048. The last 5 features consist of: the count of bounding boxes that made up this region, as well as the bounding boxes' maximum and minimum x and y coordinates relative to the total image size.

A full-image sized bounding box is used to extract the overall visual features of the image; these features are passed through two separate learnable linear layers to initialize the LSTM's hidden and cell states respectively. (These initialization steps are not featured in Fig.~\ref{fig_model_arch}.) 

After extracting the visual features from the Bottom-Up network and initializing the LSTM's hidden state and cell state (as described above), the model recursively generates a caption word-by-word until the EOS-token is produced. The process for a single step in this recursion is as follows: the LSTM's previous hidden state is passed through a single linear layer with a sigmoid activation unit to produce the \(\alpha_{t}\) weight according to equation~\ref{eq_linear_sigmoid}; this weight (and its complement) decides how much attention the model should pay to the text input and visual input respectively. The text input at each timestep is the learned word embedding for the previously generated token (or the BOS-token for the first step), while the visual input consists of the features representing the region currently indicated by the region pointer. The region pointer initially points to the first region in the requested region sequence, and is subsequently incremented each time the NEXT-token is produced by the model; if the NEXT-token is generated when no further requested regions remain, the region pointer instead points to a zero-vector region (i.e. the empty region). Once the text and visual parts of the input have been concatenated according to equation~\ref{eq_concat}, the resulting full input is passed through the LSTM module, followed by a single linear layer to produce the logits over the full training vocabulary from which the next word is sampled. This process continues until the EOS-token is produced which marks the end of the caption.

Due to the discrete-valued region pointer, the model will only receive visual input from a single region at any one timestep. However, since the LSTM's memory states are not reset between chunks, it is possible for the model to retain information about previous regions along with information about the word embeddings of previously generated words when transitioning from one chunk to the next within a single caption.

During training, the previous token refers to the ground truth token of the previous step rather than the previously generated token. The special tokens NEXT and EOS are trained in the same way as the word tokens in the vocabulary; however, the region pointer advancement follows the ground-truth NEXT-tokens only, and generating EOS early will not prevent the model from training on the remaining part of the caption.

\section{Experiments}
In our experiments, we use the Karpathy splits \cite{k_splits} of the Flickr30k \cite{flickr30k} dataset, giving us 29,000 images for training, 1014 for validation and 1000 for the test set; each image is associated with 5 human-annotated captions. In Controllable Image Captioning, a unique example is defined by an image along with a unique sequence of regions; thus, each image in the Flickr30k dataset corresponds to between 1 and 5 unique examples in our experiments (at most 1 per caption if they all use different region sequences). Thus, while the number of unique images remain the same, the number of examples in our splits become: 120,667 examples for training, 4208 for validation and 4148 for testing.

We follow the standard procedure for pre-processing captions: all captions were lower-cased and stripped of punctuation, and we replace rare words (less than 5 occurrences in the training set) by an UNK token in the training data; no words were replaced in the test set. The captions were then split into chunks based on the human-annotated entity annotations (as described in section~\ref{sec_pointer}), where each chunk ends with a visually grounded entity's associated noun phrase (or the last word of the caption respectively). During test time, each example consists of an image, a (possibly empty) sequence of regions and one or more ground-truth captions to measure success against. The Bottom-Up network \cite{bottom_up} used for the visual embeddings was trained on a custom split of the Visual Genome \cite{visual_genome} dataset (after standard pre-training on ImageNet \cite{imagenet}) to avoid an overlap between the images in the Bottom-Up net's training set and the images from our model's test set. The weights of the Bottom-Up net were frozen and remained fixed during training while the word embedding features were learned end-to-end along with the rest of the model.

We implement our model (described in section~\ref{sec_model}) using the PyTorch\footnote{https://pytorch.org/} framework. We use a batch size of 100, a learning rate of \(1e^{-5}\) and a dropout of 0.7 for both the previous word embedding and the LSTM's layer connections.\footnote{The LSTM dropout uses the method from \newcite{lstm_multi} which does not apply dropout to the recurrent connections.} To prevent overfitting, the model was evaluated on the validation set every 10 epochs, and the checkpoint with the best CIDEr \cite{cider} metric was selected. During inference on the test set for the final results, the models were prevented from generating the UNK token by setting its probability to zero. If the end-of-sequence (EOS) token was generated before all regions had been used, it was instead interpreted as a NEXT-token. If there were no more regions when a NEXT-token was generated, the visual input was referred to the empty (zero-vector) region.

All learned parameters were trained using the cross-entropy loss over the generated word sequence, to minimize the negative log probability of the ground-truth words from each caption, per equation~\ref{eq_ce}:

\begin{align}
  loss = \frac{1}{T} \sum_{t=1}^{T} -\log P(w_{t} | w_{t-1}, v_{t}, S_{t}) \label{eq_ce}
\end{align}

\noindent
where \(w_{t}\) and \(w_{t-1}\) are the current and the previous ground-truth words, \(v_{t}\) is the visual input and \(S_{t}\) is the current memory state of the model at time \(t\). All learnable weights were initialized to random uniform floats in range [-0.1, 0.1] except biases which are initialized to 0.0. The word embedding features were given a random uniform initialization in range [-1.0, 1.0].

\subsection{Ablation Tests}
The purpose of our region pointer advancement method is to enable our model to generate strongly grounded language by attending the appropriate visual region at each timestep. Meanwhile, this also allows our language model to learn valuable information about the chunk structure of the captions, which may, in itself, be a useful tool for improved language generation. To better assess the effect of appropriately timed region attention, we train an ablation version of our model (called \textit{Ours, average-pooled} in the tables), which learns the same chunk structure in the text but does not have access to the individual region features. Instead, we replace the individual region embeddings with the average-pooled features\footnote{The last five features (describing each region's bounding boxes) are replaced by zeros.} from the current example's full region sequence. Thus, the average-pooled model receives visual information for all regions in the relevant sequence, but is unable to attend to individual regions at each timestep. As with the full model, the average-pooled model is informed whether it has described the full number of regions, and it likewise receives the empty region as input when the full number of region chunks have been generated.

\section{Results and Analysis} \label{sec_results}
To test our full model's ability to learn the appropriate timing of the region pointer advancement using our proposed method, we measure how often the NEXT-token is predicted in agreement with the ground-truth from the test set when guiding the model with the correct previous word from the ground-truth data at each timestep (i.e. in the teacher-guided scenario). We find that our model generates the NEXT-token in agreement with the ground-truth to a precision of 86.55\% and a recall of 97.92\%, thus establishing that our region pointer advancement method works well in practice and as such is likely to contribute to the successful training of our model. In the rest of our experiments, the ground-truth captions are not known during inference, and the previous word refers to the model's previously sampled word.

To evaluate how our model performs on the Controllable Image Captioning task, we measure the model's performance on the standard captioning metrics: BLEU \cite{bleu}, ROUGE-L \cite{rouge}, METEOR \cite{meteor}, CIDEr \cite{cider} and SPICE \cite{spice}, using the implementation from the speaksee\footnote{https://github.com/aimagelab/speaksee} tool. The first four of these are \textit{n}-gram based metrics, with CIDEr being developed specifically for Image Captioning, while the other three have been borrowed from the machine translation and summarization fields. Among this class of metrics, CIDEr and METEOR have been associated with the highest correlation to human evaluation scores on the standard Image Captioning task \cite{ic_survey}; they provide a measure of both the fluency and content agreement of the generated captions in relation to the ground truth. Like CIDEr, SPICE was also developed as an Image Captioning metric, but instead measures the overlap between inferred scene graph tuples from the candidate and ground-truth captions, and is thus intended to measure the accuracy of the objects and their relations in the candidate captions.

We compare our results to the Show Control and Tell (SCT) model from \newcite{sct} who introduced the Controllable Image Captioning task. We compare to three versions of the SCT model: SCT (CE), SCT (CIDEr) and SCT (CIDEr NW). SCT (CE) is trained with the standard cross-entropy loss, while the other two are first trained on the cross-entropy loss and then further finetuned using Reinforcement Learning. SCT (CIDEr) is finetuned towards the CIDEr metric, while SCT (CIDEr NW) is finetuned on a combination of CIDEr and a region alignment score (defined by equation 13 from \newcite{sct}) based on word embedding similarity between nouns of corresponding caption chunks. All three SCT models use beam sampling with a beam size of 5. We present the results from our model using argmax sampling and standard cross-entropy loss training and leave further finetuning up to individual applications. For our full model we report the mean across 6 runs along with 95\% Confidence Intervals.

Table~\ref{tbl_std_metrics} shows that our full model outperforms the SCT models on all standard metrics except SPICE (with only the ROUGE-L score of SCT (CIDER NW) falling within our 95\% Confidence Interval), suggesting that our generated captions are more similar to the ground-truth captions. Interestingly, our model outperforms all SCT models on the CIDEr score, including the two SCT models that have been finetuned specifically towards this metric. All models except our average-pooled version produce shorter captions than the ground-truth's average 12.4 words, with our full model being the second closest with an average of 12.1 words. The SCT (CE) model has the shortest average caption length of 9.4 words, while the finetuned SCT (CIDER NW) increases this to 11.4.

\begin{table}
\begin{center}
\setlength{\tabcolsep}{5pt}
\caption{Standard metrics: B=BLEU, R-L=ROUGE-L, C=CIDEr, M=METEOR, S=SPICE. \textit{a} indicates mean results across 6 runs; \textit{b} indicates single-run model results from \newcite{sct}. Models above the horizontal middle line report results without finetuning.}
\label{tbl_std_metrics}
\begin{tabular}{l | cccccccc }
Model & B-1 & B-2 & B-3 & B-4 & R-L & M & C & S \\
\noalign{\smallskip}
\hline
\noalign{\smallskip}
Ours, average-pooled	&37.49 &23.18 &14.98 & 9.96& 33.38& 15.12 	 & 64.51 	& 18.70 \\
Ours, full model\textsuperscript{a}     & \textbf{41.77} & \textbf{27.14}& \textbf{18.42}&\textbf{12.83} & \textbf{38.97} & \textbf{17.33} & \textbf{87.25} 	 & 22.17 \\
\hspace{1mm} 95\% Confidence Interval			     & \(\pm 0.15\) & \(\pm 0.15\) & \(\pm 0.11\) & \(\pm 0.07\) & \(\pm 0.12\) & \(\pm 0.08\) & \(\pm 0.52\) 	 & \(\pm 0.14\) \\
\noalign{\smallskip}
SCT (CE)\textsuperscript{b} & 33.62 & 22.47 &15.68 & 11.25 & 36.86 & 15.42 	& 74.52 			& 23.45 \\
\hline
\noalign{\smallskip}
SCT (CIDEr)\textsuperscript{b} 	   & 39.26 & 25.79 & 17.58 & 12.36 & 38.84 & 16.58& 83.72 & 23.45 \\
SCT (CIDEr NW)\textsuperscript{b}   & 40.44 & 26.51 & 17.97 & 12.52 & 38.93 & 16.75& 83.99  & \textbf{23.50} \\
\hline
\end{tabular}
\end{center}
\end{table}

\begin{table}
\begin{center}
\setlength{\tabcolsep}{5pt}
\caption{Diversity statistics: Diversity = distinct captions, Novelty = captions not seen in the training set, Vocab = total number of unique words in the generated captions, Length = average number of words in the captions. \textit{a} indicates mean results across 6 runs.}
\label{tbl_div}
\begin{tabular}{l | ccr | r }
Model & Diversity \% & Novelty \% & Vocab & Length \\
\noalign{\smallskip}
\hline
\noalign{\smallskip}
Ground Truth & 99.96 & 99.70 & 4247 & 12.4 \\
\hline
\hline
\noalign{\smallskip}
Ours, average-pooled & 90.96 & 96.02 & 1042 & 12.4 \\
Ours, full model\textsuperscript{a} & \textbf{95.36} & 96.83 & \textbf{1200} & 12.1 \\
\hspace{1mm} 95\% Confidence Interval			     & \(\pm 0.32\) & \(\pm 0.13\) & \(\pm 36\) & \(\pm 0.1\) \\
\noalign{\smallskip}
SCT (CE) \cite{sct} & 86.24 & 95.98 & 935 &  9.4 \\
\hline
\noalign{\smallskip}
SCT (CIDEr) \cite{sct} & 89.98 & 97.79 & 538 & 10.9 \\
SCT (CIDEr NW) \cite{sct} & 92.06 & \textbf{98.07} & 502 & 11.4 \\
\hline
\end{tabular}
\end{center}
\end{table}

While the standard captioning metrics provide us with a measure of the fluency and content overlap with the ground-truth, they do not directly penalize a model's tendency towards repetitive captions \cite{diversity}. We expect a model with a well-functioning region pointer advancement timing to encourage more diverse and detailed captions due to better alignment with the visual input -- in comparison, if the pointer advancement lags behind the language-driven transition to the next chunk, then it would be necessary for the language model to start generating the next chunk of words without access to the relevant visual input, possibly by relying more on word co-occurrence rates in the text. To test this, we employ three metrics proposed by \newcite{diversity}: \textit{diversity} (proportion of distinct candidate captions), \textit{novelty} (proportion of candidate captions not found in the training set) and \textit{effective vocabulary size} (total number of unique words across all candidate captions).

From Table~\ref{tbl_div} we can see that all models, including our average-pooled baseline, perform well on both the Diversity and Novelty metrics, confirming that the Controllable Image Captioning setting promotes captions that are not generic or repetitive. Our full model generates the highest number of distinct captions (95.36\%), while the fully finetuned SCT (CIDEr NW) model generates the highest number of captions that were not seen in the training set (98.07\%). When it comes to generating captions with a varied vocabulary our full model has by far the largest effective vocabulary size at 1200 unique words -- ahead of SCT (CE) at 935 unique words and more than double that of the finetuned SCT models at 538 and 502 unique words respectively. The considerable decrease in SCT's vocabulary size after finetuning (despite an increase in their average caption lengths) might indicate an unwanted side-effect of CIDEr optimization, possibly encouraging a preference for common \textit{n}-grams while not sufficiently rewarding uncommon words.

Finally, from our ablation test we can tell that the average-pooled features combined with knowledge about the appropriate number of chunks is sufficient to produce acceptable results, despite using the same visual features at each timestep. However, while the average-pooled model performs acceptably, our full model is still clearly ahead on the standard captioning metrics. Thus, the results indicate that our full model does indeed learn a region pointer advancement timing that is useful for learning to generate visually grounded language.

A possible explanation for the relatively good results of the average-pooled ablation model could be that the average-pooled model learns to internally keep track of the current chunk number along with memorizing a typical sentence structure (e.g. by learning to describe people in the first chunk, followed by their attributes, and describing their activity in a later chunk). Another possibility is that it learns a type of content planning similar to older encoder-decoder models without attention, with the NEXT-token aiding in learning the sentence structure and with the additional benefit of having the empty region input indicating when to end the sequence.

\begin{figure}
  \centering
  \includegraphics[height=8cm]{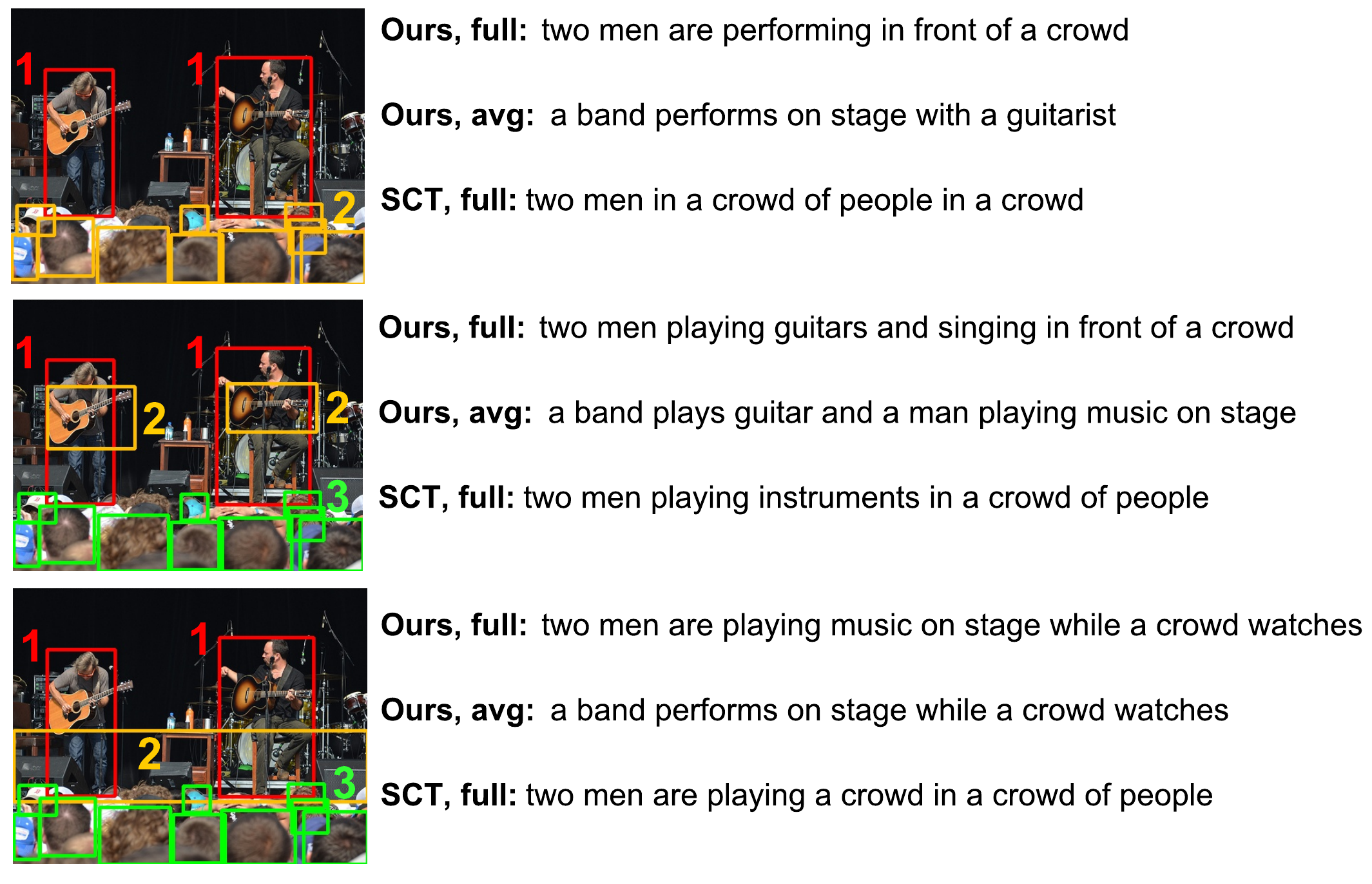}
  \caption{Variations in captions on different region sequences from the same image. \textit{SCT, full} is the fully finetuned SCT~(CIDEr NW) model from \newcite{sct}.}
   \label{fig_ex_stage}
\end{figure}

Fig~\ref{fig_ex_stage} shows the generated captions for three different region sequences on the same image. (More examples can be found in the appendix.) Overall, we found that all models were capable of producing reasonable and detailed descriptions, but that the SCT models seemed more likely to produce commonsense relationship errors (e.g. \textit{playing a crowd} from the bottom example in Fig~\ref{fig_ex_stage}). This difference can likely be explained as an effect of our model's region pointer advancement timing being strictly tied to the sentence structure, thus allowing it to attend to the next region before generating the words that tie the two regions together. In contrast, if the model was to delay its region pointer advancement with as little as a single timestep, it would need to start generating relationship words based solely on the first of the two regions with no knowledge of the second.

\section{Conclusion and Future Work}
Based on the strong correlation between sentence structure and region-related chunks in the training data's captions, we proposed a language-driven method of region pointer advancement in Controllable Image Captioning. We have implemented our proposed method in a Controllable Image Captioning model where it demonstrates a precision of 86.55\% and a recall of 97.92\%. Our full model outperforms the current state of the art model on the standard metrics, including CIDEr, despite using only the cross-entropy loss whereas the current state-of-the-art relies on finetuning towards the CIDEr metric.

Additionally, we find that our model has an effective vocabulary size that is more than double that of the current state-of-the-art, suggesting that our model is more capable of learning and generating uncommon words.

We have demonstrated that our method for region pointer advancement works well in the vision-to-text context. However, its implementation could be applied to any sequence-to-sequence tasks where structural chunks in the input data (e.g. image regions) can be related to structural chunks in the output (e.g. natural language sentence chunks); some possible applications would be Speech-to-Text, Machine Translation or the standard Image Captioning task when combined with a region selection and sorting mechanism.

For the task of Controllable Image Captioning, we would encourage future work to consider complementary metrics such as caption diversity and effective vocabulary size (alongside the standard captioning metrics) to better understand a model's capacity to generate unique descriptions for each unique input. Additional metrics to specifically measure the adherence to an ordered region sequence would be welcome.

\section*{Acknowledgements}
This research was conducted with the financial support of Science Foundation Ireland under Grant Agreement No. 17/RC-PHD/3488 at the ADAPT SFI Research Centre at Technological University Dublin.  The ADAPT SFI Centre for Digital Media Technology is funded by Science Foundation Ireland through the SFI Research Centres Programme and is co-funded under the European Regional Development Fund (ERDF) through Grant \# 13/RC/2106.

\bibliographystyle{coling}
\bibliography{controllable}

\vspace*{\fill}

\pagebreak
\appendix
\section*{Appendix. Additional Caption Examples.}

\begin{figure}[!h]
\centering
\includegraphics[height=17cm]{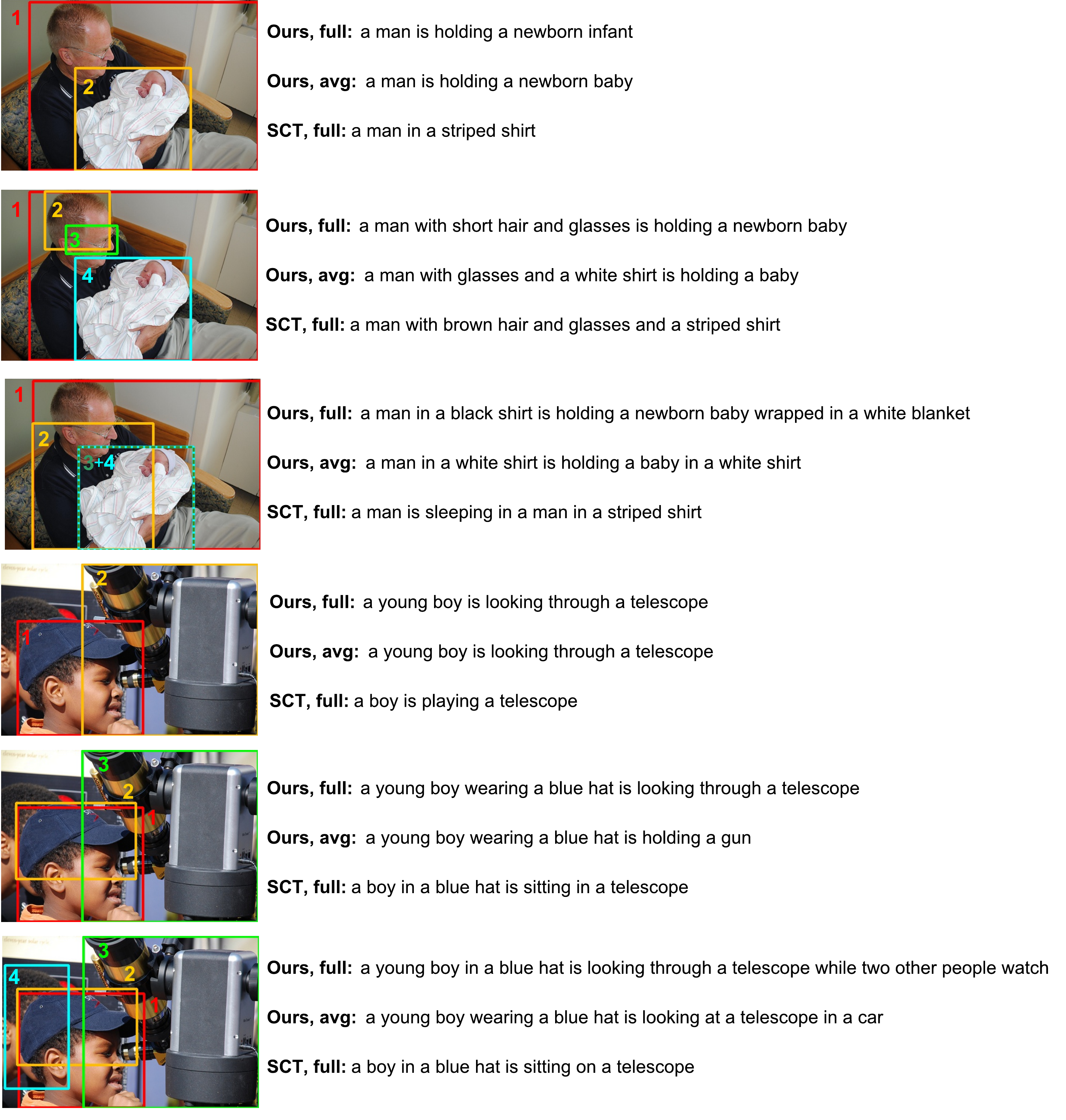}
\caption{Additional examples of generated captions. \textit{SCT, full} is the fully finetuned SCT~(CIDEr NW) model referenced in Section~\ref{sec_results}.}
\label{fig_appendix_1}
\end{figure}

\begin{figure}
\centering
\includegraphics[height=17cm]{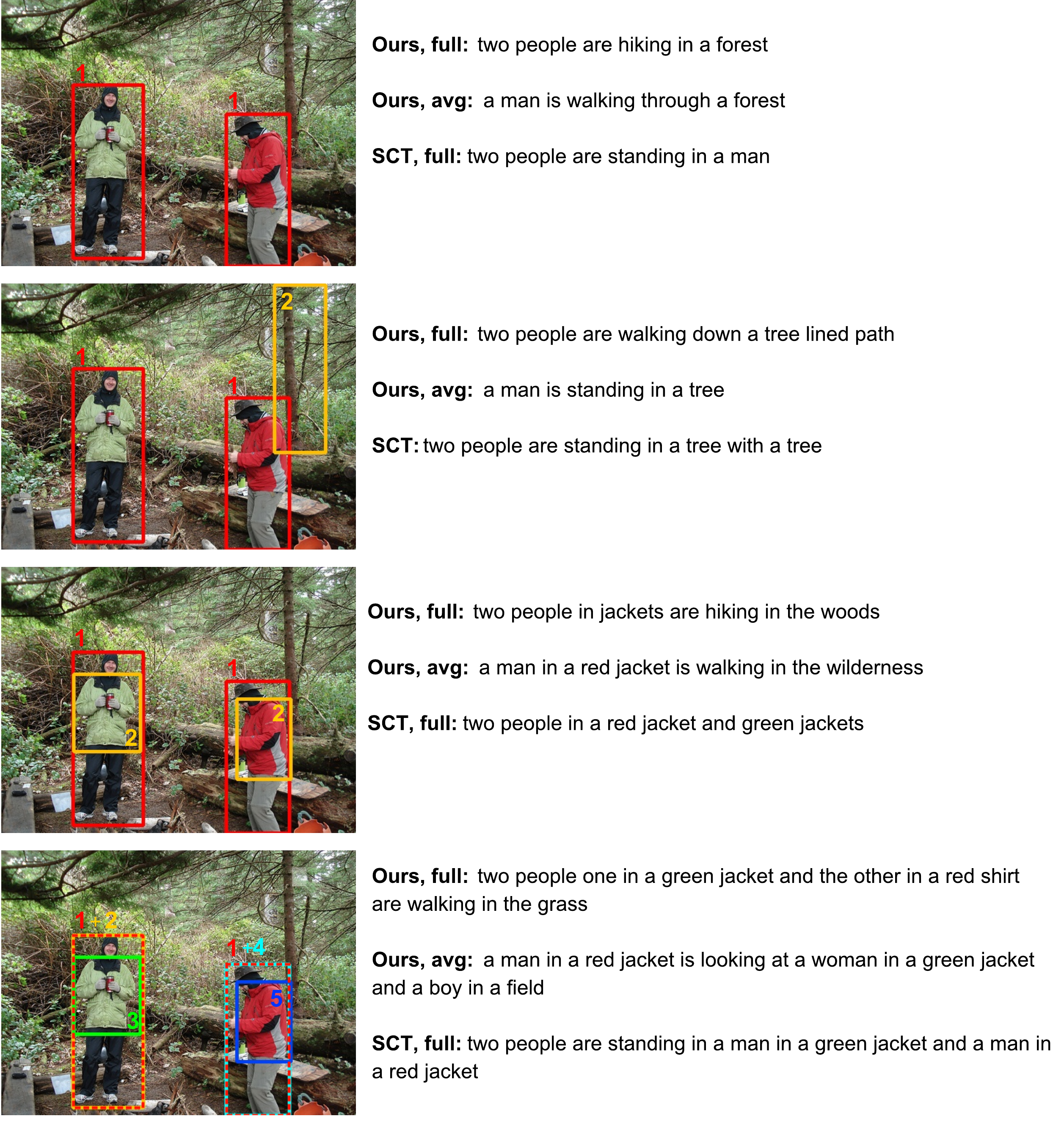}
\caption{Additional examples of generated captions. \textit{SCT, full} is the fully finetuned SCT~(CIDEr NW) model referenced in Section~\ref{sec_results}.}
\label{fig_appendix_2}
\end{figure}

\begin{figure}
\centering
\includegraphics[height=17cm]{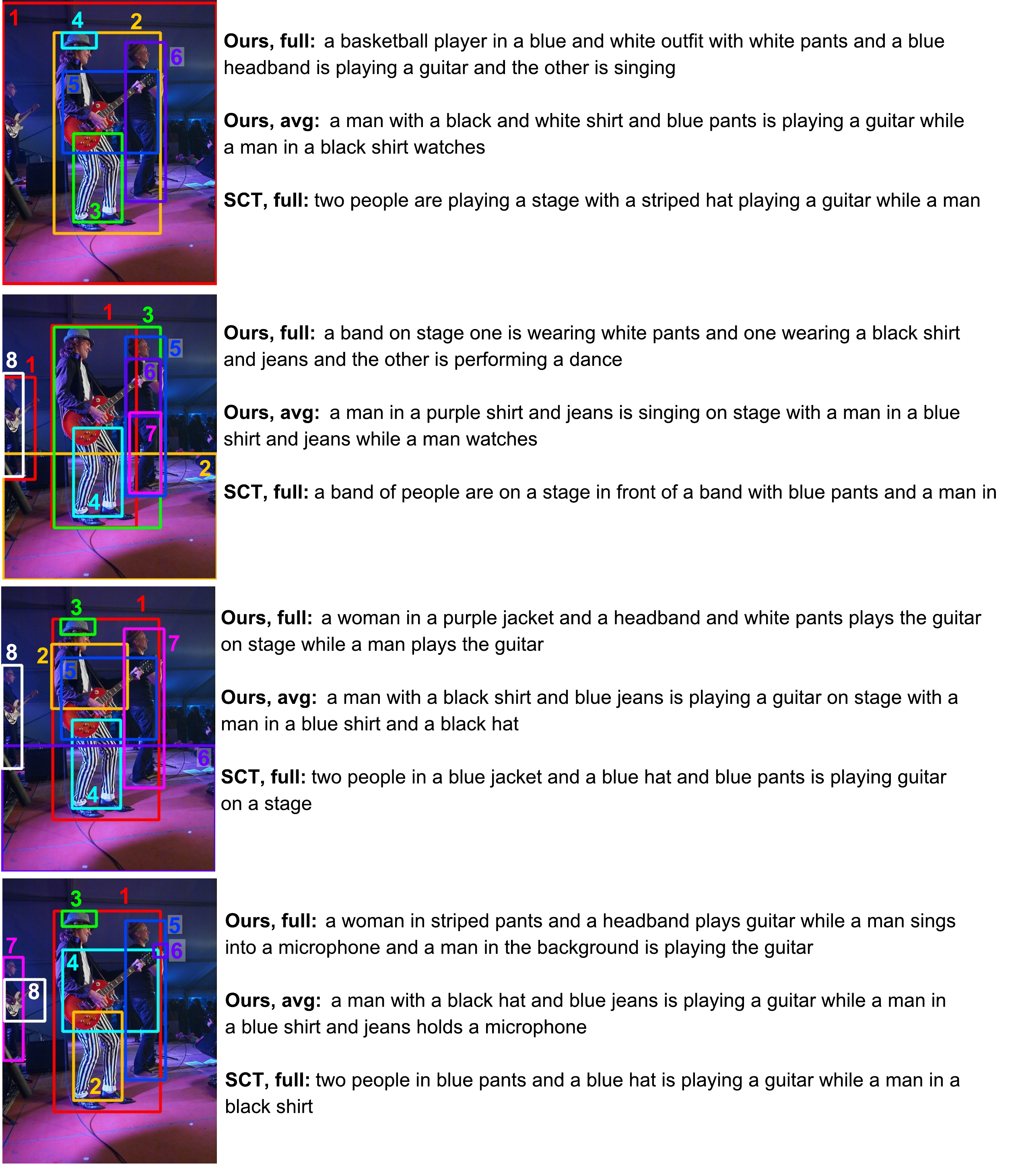}
\caption{Additional examples of generated captions. \textit{SCT, full} is the fully finetuned SCT~(CIDEr NW) model referenced in Section~\ref{sec_results}.}
\label{fig_appendix_3}
\end{figure}

\end{document}